\documentclass[11pt]{article}

\usepackage[preprint]{acl}

\usepackage{times}
\usepackage{latexsym}

\usepackage{caption}
\usepackage{booktabs}
\usepackage{tabularx}
\usepackage{array}
\usepackage{makecell}
\usepackage{xcolor}

\newcolumntype{Y}{>{\raggedright\arraybackslash}X}

\usepackage[T1]{fontenc}

\usepackage[utf8]{inputenc}

\usepackage{microtype}
\usepackage{inconsolata}

\usepackage{graphicx}

\usepackage{amsmath}
\usepackage{amssymb}

\usepackage{tikz}
\usetikzlibrary{shapes.geometric, arrows.meta, positioning, calc, fit, backgrounds, decorations.pathreplacing}

\definecolor{dkgreen}{HTML}{3B6B5C}
\definecolor{coral}{HTML}{C0392B}
\definecolor{cream}{HTML}{FDF3DC}
\definecolor{ltgray}{HTML}{EAF0F0}
\definecolor{medgray}{HTML}{90A4AE}
\definecolor{txtdark}{HTML}{2C3E50}

\title{Affording Process Auditability with QualAnalyzer: An Atomistic LLM Analysis Tool for Qualitative Research}

\author{
Max Hao Lu \\
Harvard University / United States \\
\texttt{maxlu@fas.harvard.edu}
\And
Ryan Ellegood \\
Harvard University / United States \\
\texttt{ryanellegood@gse.harvard.edu}
\AND
Rony Rodriguez-Ramirez \\
Harvard University / United States\\
\texttt{rrodriguezramirez@g.harvard.edu}
\And
Sophia Blumert \\
Harvard University / United States\\
\texttt{sophia\_blumert@gse.harvard.edu}
}

\begin{document}
\maketitle

\begin{abstract}
Large language models are increasingly used for qualitative data analysis, but many workflows obscure how analytic conclusions are produced. We present QualAnalyzer, an open-source Chrome extension for Google Workspace that supports atomistic LLM analysis by processing each data segment independently and preserving the prompt, input, and output for every unit\footnote{The Chrome extension is available from the \href{https://chromewebstore.google.com/detail/llm-qualanalyzer/ndadgbdohohicninakmhdnllknkbjggj}{Chrome Web Store}, and the source code repository is available on \href{https://github.com/maxlvhao/llm_qual_analyzer}{GitHub}.}. Through two case studies---holistic essay scoring and deductive thematic coding of interview transcripts---we show that this approach creates a legible audit trail and helps researchers investigate systematic differences between LLM and human judgments. We argue that process auditability is essential for making LLM-assisted qualitative research more transparent and methodologically robust.
\end{abstract}

\section{Introduction}
Large language models (LLMs) are increasingly used across the qualitative research workflow, from transcription and data preparation to coding and synthesis \citep{brailas2025artificial, christou2023use}. Their appeal is straightforward: they can process large volumes of text quickly and produce plausible analytical outputs with little setup. Yet many current LLM workflows remain difficult to audit. When researchers ask a chatbot to summarize, code, or interpret a dataset, the resulting output often appears without a clear record of what evidence was considered, how instructions were applied, or how conclusions were generated \citep{jones2025generative}.

This opacity matters because qualitative rigor depends not only on the final interpretation but also on the traceability of the analytic process. Human qualitative coding is not fully transparent either, but it is typically embedded in procedures that make interpretation accountable: researchers work from shared codebooks, compare judgments, document disagreements, and revise analytic decisions over time \citep{williams2019art}. By contrast, many LLM-based workflows provide neither a systematic audit trail nor an easy way for research teams to inspect, compare, and iteratively refine model behavior across an entire dataset.

Existing tools offer only partial solutions. Off-the-shelf chat interfaces obscure analytical steps, qualitative data analysis platforms provide limited support for iterative researcher-defined LLM coding workflows, and scripted API pipelines are often inaccessible to non-programmers. In our own large-scale qualitative work, we found no tool that supported segment-level LLM analysis at scale while preserving transparency, iteration, and collaborative oversight. 

We present \textit{QualAnalyzer}, an open-source tool that supports what we call \textit{atomistic LLM analysis}: processing each data segment independently, preserving the prompt and output for every unit, and making those records available for inspection and comparison. Built as a Chrome extension for Google Workspace, QualAnalyzer supports segment-level analysis, prompt iteration, audit trails, and reliability checks in a collaborative environment familiar to many qualitative researchers. Through two case studies—holistic essay scoring and deductive coding of interview transcripts—we show how this workflow makes LLM-assisted analysis more auditable and how preserved outputs help researchers investigate discrepancies between human and model judgment.

Taken together, the paper contributes a methodological argument for process auditability in LLM-assisted qualitative research and a practical system for supporting it in collaborative workflows.

\section{From Holistic to Segment-Level LLM Analysis}

\begin{figure*}[!t]
\centering

\includegraphics[width=\textwidth]{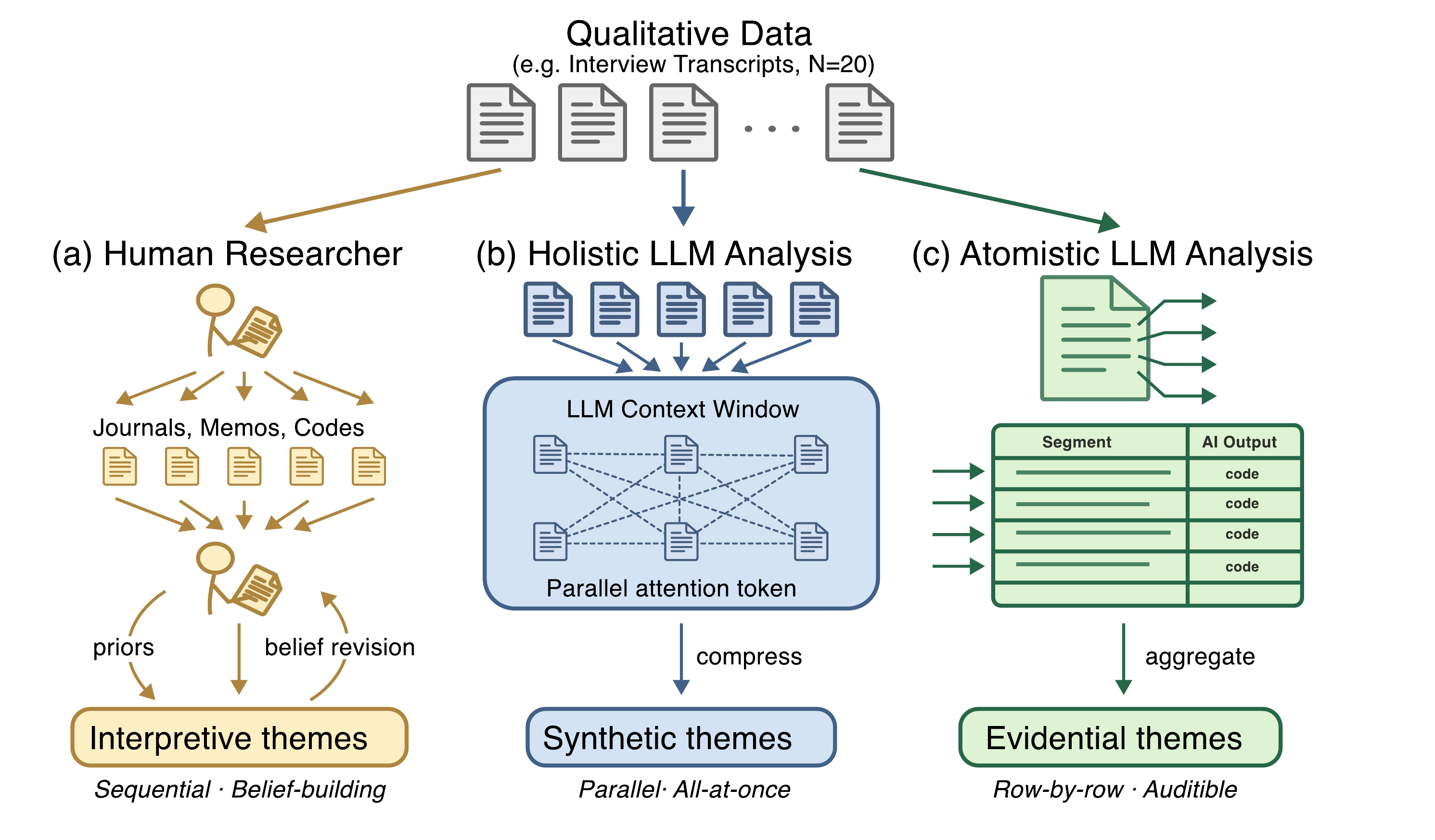}
\caption{Three paradigms for LLM-assisted qualitative analysis of interview data. (a) The traditional human researcher approach processes documents sequentially. (b) Holistic LLM analysis loads all documents into a single context window for parallel processing. (c) The proposed atomistic LLM analysis processes each document independently, row-by-row, and aggregates AI-generated codes into evidential themes.}
\label{fig:ThreeParadigms}

\vspace{0.6em}

\captionsetup{type=table}
\caption{Three-way comparison framework for AI-assisted qualitative analysis. Each dimension contrasts human analysis with holistic (full-context) and segmented (segment-level) LLM workflows, grounded in prior literature}
\label{tab:comparison-framework}

\small
\setlength{\tabcolsep}{4pt}
\renewcommand{\arraystretch}{1.15}
\begin{tabularx}{\textwidth}{>{\raggedright\arraybackslash}p{0.16\textwidth} Y Y Y}
\toprule
\textbf{Dimension} &
\textbf{Human Researcher} &
\textbf{Holistic LLM Analysis} &
\textbf{Atomistic LLM Analysis} \\
\midrule

\textbf{How attention is distributed} &
Sequential and selective; shaped by fatigue, interest, and prior beliefs \citep{belur2021interrater}. &
Parallel via self-attention across tokens; subject to positional biases (``lost in the middle'') in long contexts \citep{liu2024lost}. &
Uniform and controlled; each segment receives comparable computational attention without positional disadvantage. \\

\textbf{How understanding is built} &
Iterative belief formation via sequential reading; constant comparison in grounded theory. \citep{charmaz2017constructivist,glaser2017discovery} &
Single-pass parallel processing; relationships computed via attention within transformer layers. \citep{vaswani2017attention} &
Incremental per-unit analysis, avoids whole-text aggregation bias but can miss holistic insights. \\

\textbf{Cross-document pattern detection} &
Limited by serial memory; supported by memos and coding systems. \citep{saldana2025coding} &
Strong in principle (cross-token linking), but constrained by long-context failures. \citep{liu2024lost} &
Weaker by default, each analysis targets a single atomic unit, cross-document patterns require post-coding aggregation or cross-references.\\

\textbf{Auditability / process transparency} &
Moderate; depends on reflexive documentation and analytic practices. \citep{o2020intercoder} &
Low; conversational querying yields opaque reasoning traces.&
High; per-unit outputs + recorded prompts produce inspectable audit trails.\\

\textbf{Hallucination / fabrication risk} &
Low for fabrication; risks of projection/over-interpretation.&
Higher; LLM hallucination widely documented. \citep{huang2025survey} &
Lower per unit; shorter context reduces room for fabrication.\\

\bottomrule
\end{tabularx}

\end{figure*}
Qualitative analysis has long valued reflexivity, interpretive accountability, and close engagement with data. The rapid adoption of LLMs for analytic work risks weakening those qualities when model outputs are produced without a clear record of how they were generated. In qualitative practice, a researcher reads transcripts, marks passages, assigns codes, writes memos, and gradually builds an interpretation. The process can be slow, but that slowness is part of the rigor: each transcript is read through the evolving lens of what has already been seen, so understanding accumulates sequentially \citep{charmaz2017constructivist, glaser2017discovery}. The researcher's closeness to the data is therefore not a byproduct of the method but part of the method itself.

\textbf{Limitations of the Holistic LLM Analysis Paradigm.} With the rise of large language models, a faster alternative has gained traction. Researchers upload transcripts into a chat interface, pose analytic questions, and receive thematic summaries or code suggestions in return. We call this \textit{holistic LLM analysis} (Figure~\ref{fig:ThreeParadigms}): the model ingests the available data and produces an output in a single pass. For exploratory tasks such as generating an initial overview or checking whether a dataset contains a particular topic, this can be highly productive. However, holistic analysis is poorly suited to settings where researchers need to inspect how segment-level judgments were made. Because the model produces a single integrated response over a large context, it is often difficult to trace which passages informed a given conclusion and which were ignored. Long-context processing is also uneven: models tend to weight the beginning and end of long inputs more than the middle, increasing the risk of omissions and hallucinated findings \citep{liu2024lost}.

LLMs’ opaque reasoning, combined with the interpretive ambiguity of qualitative coding, means this is rarely a one-prompt task. Constructs such as ``critical thinking'' or ``growth mindset'' are difficult even for human coders to apply consistently, and translating those judgments into reliable prompts requires iteration and validation. We therefore use \textit{process auditability} to mean the extent to which an analytic conclusion can be traced: which data were analyzed, what instructions were applied, what outputs were produced for each unit, and how those outputs were combined into findings. Audit trails, memos, and reflexive documentation have long been central to qualitative rigor \citep{o2020intercoder, saldana2025coding}; with LLMs, the need is more acute because one-shot summaries often preserve no segment-level record of what evidence was used or omitted.

\textbf{Existing tools and their limitations.} CAQDAS platforms such as ATLAS.ti, MAXQDA, and NVivo now include LLM-based features such as summaries and code suggestions, but these features largely support emergent or in vivo coding rather than researcher-defined prompt iteration and reliability checking across a full dataset \citep{mortelmans2025doing}. Scripted Python or R pipelines can preserve segment-level transparency, but they place the logic of analysis in code that many collaborators cannot easily inspect or revise. The gap, then, is practical as much as technical: qualitative teams need infrastructure for segment-level LLM analysis that is visible, iterative, and usable without programming.

\textbf{Atomistic LLM analysis.} We propose \textit{atomistic LLM analysis} as a complement to holistic analysis. In this approach, each document or segment is processed independently with a standardized prompt, and each segment's output is recorded separately. More precisely, where holistic analysis produces a single undifferentiated output over the entire corpus,
\begin{equation}
  Y^{\mathrm{hol}}
  = \mathcal{M}_{\psi}\!\bigl(\tau(D_1 \,\|\, D_2 \,\|\, \cdots \,\|\, D_N)\bigr),
\end{equation}
atomistic analysis processes each segment independently and preserves a one-to-one map between inputs and outputs:
\begin{equation}
\begin{split}
  y_{i,j} &= \mathcal{M}_{\psi}\!\bigl(\tau(s_{i,j},\, x_{i,1},\, \dots,\, x_{i,K-1})\bigr),\\
  &\quad \forall\; i \in \{1, \dots, N\},\; j \in \{1, \dots, n_i\},
\end{split}
\end{equation}
where $\mathcal{M}_{\psi}$ is the LLM parameterized by configuration $\psi$ (model, temperature, max tokens), $\tau$ is the prompt template, $s_{i,j}$ is the $j$-th segment of document $D_i$, and $x_{i,1}, \dots, x_{i,K-1}$ are optional context fields. This avoids position effects within a batch and preserves a segment-level record of what the model received and produced. Full formal definitions are provided in a companion document.\footnote{Formal definitions are available on OSF: \href{https://osf.io/jzg9t}{osf.io/jzg9t}.} Figure~\ref{fig:ThreeParadigms} illustrates the approach, and Table~\ref{tab:comparison-framework} compares human, holistic LLM, and atomistic LLM analysis across key dimensions.

The paper mainly contributes to frame segment-level LLM processing as a distinct methodological paradigm and to make that paradigm accessible to non-programmer research teams through an interface designed for inspection, iteration, and collaborative use.

\section{Design Principles and Solution Overview}
\begin{figure*}[t]
\centering
\includegraphics[width=\textwidth,height=0.55\textheight,keepaspectratio]{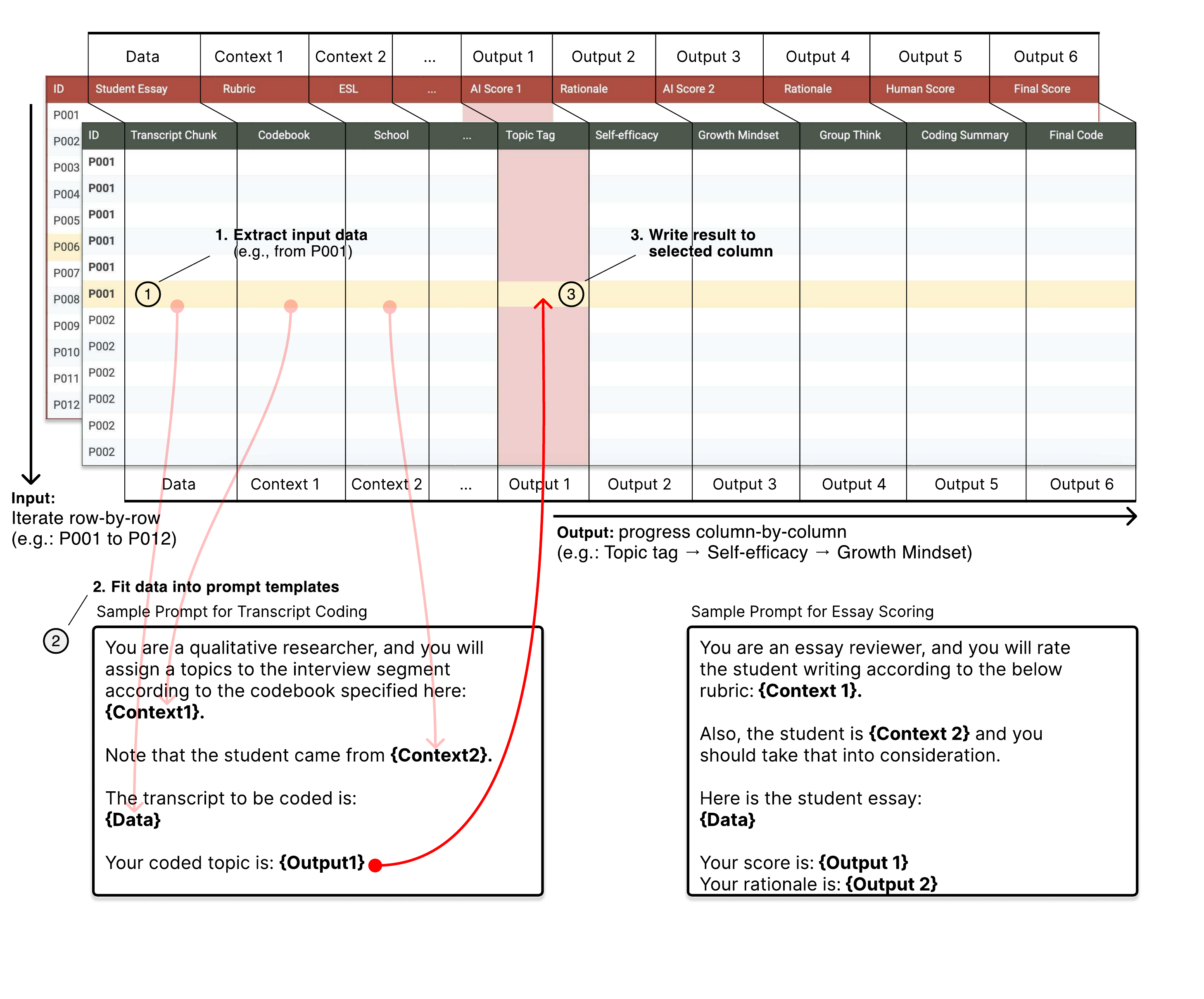}
\caption{Atomistic analysis workflow in QualAnalyzer. Each row represents an analytic unit (Data) with optional contextual fields (Context 1--2). QualAnalyzer concatenates these fields into a predefined prompt template, runs the model iteratively (row-by-row and output-by-output), and writes responses back into structured Output columns (e.g., score and rationale). This tabular schema supports reproducible runs, side-by-side model comparisons, and downstream validation. The two header styles illustrate supported input formats: (top, red) one row per complete document (e.g., a student essay) and (bottom, gray) chunked analysis of longer sources, where a single record is segmented and each transcript chunk is coded independently.}
\label{fig:SheetBreakdown}
\end{figure*}

The problems identified above suggest that infrastructure for LLM-assisted qualitative analysis should do four things well: preserve segment-level visibility into model behavior, make prompt iteration practical across a full dataset, support diagnosis through stored outputs and agreement metrics, and remain accessible to collaborative research teams without requiring programming expertise. It should also minimize unnecessary data transfer and remain extensible as research needs evolve.

QualAnalyzer was built around these requirements as an open-source Chrome extension for Google Workspace. Google Sheets serves as the primary analysis workspace: each row is a data segment, columns store context and model outputs, and adjacent columns support direct comparison across prompt versions, coding rounds, and models (Figure~\ref{fig:SheetBreakdown}). Google Docs supports data preparation and segmentation before analysis. By building on familiar collaborative tools, QualAnalyzer turns process auditability into an ordinary part of the workflow rather than a separate technical exercise.

\section{Tool Design}

\subsection{System Architecture}

\begin{figure*}
\centering
\includegraphics[width=\textwidth]{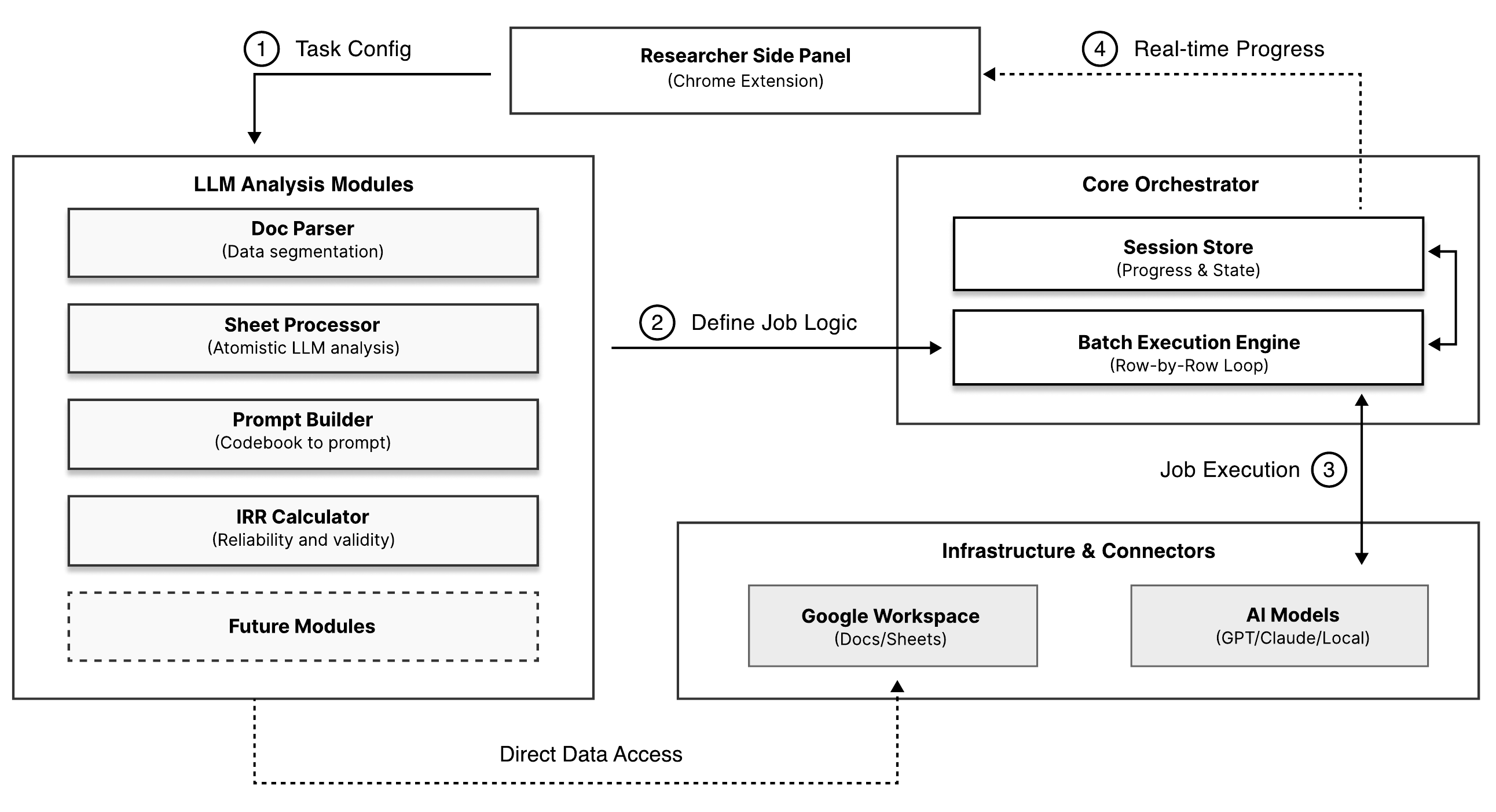}
\caption{System architecture of QualAnalyzer. The researcher configures tasks through a Chrome Extension side panel~(1). Analysis modules define job logic and pass it to the orchestrator~(2), which executes jobs by reading data from Google Sheets and sending requests to the selected LLM~(3). Progress and results return to the side panel in real time~(4).}
\label{fig:Architecture}
\end{figure*}

QualAnalyzer comprises four layers: a researcher side panel, analysis modules, a core orchestrator, and an infrastructure layer for Google Workspace and LLM access. Together, these layers are designed not only to execute LLM-based analysis, but to preserve the information needed to inspect, compare, and revisit each analytic run.

The \textbf{Researcher Side Panel} is a Chrome Extension that opens alongside Google Docs or Sheets. It lets researchers configure modules, set parameters, and monitor progress without leaving their workspace.

The \textbf{LLM Analysis Modules} implement distinct workflow stages. The \textit{Doc Parser} segments documents into analytic units (e.g., paragraph, speaker turn, delimiter, or sentence count) and writes them as rows in a Google Sheet. The \textit{Prompt Builder} converts a codebook into a prompt template. The \textit{Sheet Processor} performs atomistic analysis by processing rows sequentially, sending each segment to the selected LLM with a standardized prompt. The \textit{IRR Calculator} computes inter-rater reliability metrics (e.g., Cohen's kappa, percent agreement) across output columns, supporting comparisons across LLM runs, human and LLM codes, or prompt versions. Modules can be used independently or as a pipeline, and the system can be extended with additional modules such as cross-case synthesis.

The \textbf{Core Orchestrator} manages execution. Its Batch Execution Engine reads rows, assembles prompts, calls the LLM API, and writes responses back to the sheet. The Session Store tracks state and progress, supporting live monitoring and job resumption after interruption.

The \textbf{Infrastructure Layer} connects to Google Workspace for data access and to LLM providers for model calls. It currently supports OpenAI, Anthropic, and local models via Ollama. Researchers supply their own API keys and choose the model, preserving control over model version for reproducibility.

\subsection{User Interface and Interaction Design}

%\begin{figure} 
%\centering
%\includegraphics[width=\columnwidth]{figs/UI_WalkThrough.pdf}
%\caption{QualAnalyzer interface walkthrough. (A)~Configuring providers: connecting a Google account and adding LLM providers with explicit model selection. (B)~Selecting a function: choosing an analysis module and specifying job details. (C)~Tracking progress: monitoring jobs and auditing job history.}
%\label{fig:UI_WalkThrough}
%\end{figure}

Researchers interact with QualAnalyzer through a side panel embedded in Google Docs or Sheets. From this panel, they configure providers and models, select an analysis module, specify job parameters, and monitor execution. During processing, the interface reports job progress; after completion, it preserves a history of runs including the module used, model, prompt, source, row range, and timestamp.

The interface is intentionally minimal. Rather than replacing Docs or Sheets, it keeps controls in one place while leaving data and outputs in the shared workspace, where collaborators can inspect results directly. This design choice matters methodologically as well as practically: it makes model behavior visible in the same environment where teams already review data, discuss disagreements, and revise analytic decisions.

\section{Case Studies}
 
We demonstrate QualAnalyzer through two case studies that differ in data type, coding logic, and evaluative framework to examine what preserved segment-level outputs enable in practice.

The first case uses a subset of the Automated Student Assessment Prize (ASAP) dataset \citep{boulanger2022asap}, a public collection of student essays written to eight prompts and scored by trained raters using a holistic rubric. We focus on Essay Set 1, a Grade 8 persuasive writing task scored on a six-point scale. 

The second case uses semi-structured interview transcripts from a study on qualitative data curation practices \citep{mannheimer2022data}, archived in the Qualitative Data Repository \citep{mannheimer2023interviews}. Participants from three groups (big social researchers, qualitative researchers, and data curators) were interviewed about epistemological, ethical, and legal issues in data reuse. Here, the LLM conducts deductive thematic coding guided by a researcher-developed codebook. Across both studies, we analyze \textit{process auditability} at each workflow stage.
 
\subsection{Case Study 1: ASAP Essay Scoring}
 
The ASAP dataset includes a tabular file in which each essay occupies a single row with human scores in adjacent columns. Because this structure was already available, no document parsing or chunking was required; the data was imported directly into Google Sheets, and prompt construction served as the starting point.

The scoring rubric provided with the ASAP dataset was used as the source material for the \textit{Prompt Builder}. Guided step by step, the rubric was translated into a machine-ready prompt that defined the construct of \textit{development}, specified score-level classification rules, identified observable indicators, and returned both a numeric score and a written justification with quoted evidence from the essay. Scored anchor papers from the ASAP assessment materials were incorporated as few-shot examples, and the completed prompt was saved to the \textit{prompt library} for inspection, versioning, and iterative refinement. The full prompt is available in the supplementary materials.

Two atomistic analysis jobs were then run in the \textit{Sheet Processor} using the saved prompt, with essay text as input and outputs written to separate columns. Both jobs used GPT~5.2. This dual-pass design mirrors the double-scoring structure of the original ASAP framework and tests within-model consistency across repeated administrations of the same prompt as suggested in \citet{jung2025model}.

Each output contains a numeric score and a written justification with quoted evidence from the essay. Spreadsheet formulas extract the numeric score, combine the two passes into a tabulated LLM score, and compute differences from the human ratings. Because the justifications remain alongside the extracted values, a reviewer can inspect not only the assigned score but the reasoning that produced it. Across the two runs, the numeric scores were highly consistent, suggesting that the prompt functioned as a stable scoring instrument.

The resulting record includes the original essay text, human scores, both LLM outputs, extracted scores, and job metadata such as model, prompt, row range, and timestamp. Together, these materials make each scoring decision traceable from rubric to prompt to model output, illustrating how atomistic analysis can function as an auditable scoring workflow rather than as a one-shot prediction system.
 
\subsection{Case Study 2: Qualitative Interview Coding}
 
The second case study applies QualAnalyzer to semi-structured interview transcripts from \citet{mannheimer2022data}, a study of data curation practices across big social researchers, qualitative researchers, and data curators. Twenty-eight of thirty participants consented to share transcripts, and twenty-five were available in the published dataset \citep{mannheimer2023interviews}. The original study used a deductive codebook organized around six parent codes.

Unlike the ASAP dataset, these materials required preparation before analysis. Each transcript was imported into Google Drive and processed through the \textit{Doc Parser} in ``Entire File'' mode, producing a Google Sheet in which each transcript occupied one row with participant IDs preserved.

We selected three sub-codes for analysis: \textit{Documentation and Metadata}, \textit{IRB}, and \textit{Consent Language and Procedures}. Because the published codebook listed labels but not full operational definitions, prompt construction required consulting the dissertation to recover traceable definitions for these constructs. We limited the analysis to sub-codes for which this documentation was sufficiently specific, rather than introducing unsupported interpretive definitions ourselves.

Each prompt produced two outputs: a binary Present/Absent classification and a count of mentions with quoted supporting passages. This structure makes coding decisions auditable at the passage level by preserving both the classification and the textual evidence used to support it.
 
As in Case Study~1, two passes were run for each sub-code using GPT~5.2, with outputs written to separate columns. Three transcripts (DC07, DC08, DC09) contained missing data and were excluded, leaving twenty-three transcripts per sub-code. The \textit{IRR Calculator} was then used to compare the two passes. On the binary Present/Absent classification, agreement was perfect across all three codes (Cohen's kappa $k= 1.000$, $n = 23$). Mention counts showed exact agreement in 83--87\% of transcripts, with mean differences of 0.13 to 0.22 mentions per transcript; all disagreements were by a single mention.

Comparison with the published codebook \citep{mannheimer2023interviews} showed close alignment on binary classification but a systematic excess in mention counts. The LLM identified 10 transcripts containing \textit{Documentation and Metadata} (codebook: 9), 21 containing \textit{IRB} (codebook: 22), and 15 containing \textit{Consent Language and Procedures} (codebook: 15), despite operating on twenty-three transcripts rather than thirty. For reference counts, however, it identified 17--19 mentions for \textit{Documentation and Metadata} (codebook: 11), 43--44 for \textit{IRB} (codebook: 33), and 40--41 for \textit{Consent Language and Procedures} (codebook: 28).

The original paper does not resolve why this divergence occurs, but QualAnalyzer makes it directly investigable. Because quoted supporting passages are preserved alongside each count, researchers can review the additional mentions, determine whether they reflect legitimate references or overcounting, revise the prompt, and compare revised runs side by side. More importantly, this case shows that the value of process auditability is not limited to documenting outputs after the fact; it also supports iterative diagnosis when LLM judgments diverge from human interpretation.

\subsection{Cross-Case Reflection}
 
Across both case studies, the workflow remained the same: a rubric or codebook was translated into a prompt, the prompt was applied row by row, and repeated runs provided a basis for checking consistency. What differed was the role of the audit trail. In Case Study~1, it functioned primarily as confirmatory documentation of a stable scoring process. In Case Study~2, it functioned as a diagnostic resource, making a discrepancy in mention counts visible, reviewable, and revisable. Taken together, the cases suggest that process auditability is valuable not only for documenting analytic decisions, but also for identifying when those decisions require further scrutiny.

Both cases also show that prompt quality depends heavily on source materials. The ASAP rubric could be translated with little additional interpretation, whereas the Mannheimer codebook required supplementation from the dissertation before prompts could be constructed traceably. This suggests that LLM-assisted qualitative coding places particular demands on codebook specificity.

\section{Discussion}

This paper introduces atomistic LLM analysis as a distinct paradigm for qualitative research and presents QualAnalyzer as one way of operationalizing that paradigm for teams without programming expertise. The core issue is a methodological tension: LLMs offer powerful language-processing capabilities and ease of use, but their analytic process is often opaque, concealing the interpretive steps that qualitative researchers have long insisted must remain traceable. Atomistic analysis addresses this tension by preserving the segment-level input, instruction, and output associated with each analytic judgment.

A central design decision was to embed the tool within researchers' existing workflows rather than asking researchers to migrate into a new environment. By operating inside Google Workspace, QualAnalyzer keeps data, outputs, and collaboration in a space teams already use, while the spreadsheet format makes each step of the model's processing visible in a structure familiar to most researchers. This matters not only for usability but for oversight: when every LLM output occupies a cell alongside the input that produced it, collaborative review becomes a natural extension of ordinary research practice rather than a separate auditing task.

The two case studies illustrate that this paradigm generalizes across research contexts. In automated essay scoring, the audit trail functioned as confirmatory documentation of a stable instrument. In deductive interview coding, the same infrastructure served a diagnostic purpose, surfacing a systematic divergence in mention counts that the preserved outputs made investigable. In both cases, it is useful to think of the LLM as a highly capable research assistant — one whose contributions become productive only when embedded in protocols that make its reasoning transparent and subject to the same scrutiny applied to any other member of the research team.

QualAnalyzer assembles existing capabilities---API calls, spreadsheet integration, and reliability computation---into a workflow designed around a specific methodological need. We view this as a contribution precisely because the barrier to transparent LLM use in qualitative research has been practical rather than technical: the underlying components already exist, but they have not been organized in a way that supports the iterative, collaborative, and auditable workflows qualitative research demands. The larger point is therefore not only that this particular tool is useful, but that transparent LLM-assisted qualitative analysis requires infrastructure explicitly designed for inspection and revision rather than only for output generation. In this sense, the value of atomistic analysis lies less in automation alone than in making model-assisted interpretation available for inspection, contestation, and revision.

Finally, atomistic analysis is best understood as a complement to holistic analysis, not a replacement for it. The two paradigms serve different stages of interpretive work: holistic interaction supports familiarization and synthesis, while atomistic processing supports segment-level judgments that benefit from standardized prompts and preserved outputs. One can envision hybrid workflows that traverse between the two — generating cross-document insights holistically while maintaining the systematic rigor that atomistic analysis affords. Across all such configurations, however, the researcher remains the primary analytical agent. The LLM produces structured traces; the researcher decides how those traces should be interpreted and whether they support a defensible finding.

\section*{Limitations}
 
QualAnalyzer provides infrastructure for structuring, documenting, and assessing LLM-assisted qualitative analysis within an environment accessible to researchers without programming expertise. This accessibility is by design, but it introduces a risk. The tool makes the analytical process \textit{visible}; it cannot ensure the researcher \textit{engages} with what is visible. A complete audit trail (preserved prompts, raw outputs, extracted scores, reliability statistics) can be produced without the researcher ever inspecting the model's justifications, reviewing flagged passages, or questioning why the LLM coded a particular segment the way it did. Transparency in this sense is passive: the infrastructure for scrutiny is present, but the scrutiny itself remains a human responsibility. Future iterations of the tool should explore ways to encourage active engagement at key decision points rather than allowing the pipeline to run from configuration to results without intervention.
 
The \textit{atomistic} approach that enables this auditability is itself a constraint on the kinds of analysis the tool can support. Because each segment receives an independent pass through the model, coding schemes that require synthesizing across multiple segments to produce a single judgment are not currently accommodated. Holistic codes, where a researcher reads an entire transcript and assigns an overall characterization based on patterns distributed throughout the document, depend on exactly the kind of cross-segment reasoning that atomistic analysis deliberately avoids. The ``Entire File'' chunking mode used in Case Study~2 partially addresses this by treating the full transcript as a single segment, but this workaround reintroduces the long-context processing concerns discussed in Section~2. Supporting analyses that require structured cross-segment synthesis while preserving per-unit auditability remains an open design challenge.
 
Even within the analyses the tool does support, a further distinction applies: \textit{process auditability} documents how results were produced, but it does not establish whether those results are correct. As the case studies illustrate, high within-model consistency does not preclude systematic divergence from human judgment. Consistency does not automatically imply validity. Reliable metrics indicate that the analytical instrument is stable; they do not indicate that it measures what the researcher intends it to measure. The tool can structure the process, record the decisions, and flag inconsistencies, but interpreting those inconsistencies, revising the prompt, and determining whether the coding aligns with the research questions remain tasks that require a human researcher with substantive familiarity with the data.

\section*{Acknowledgments}
We thank the Kern Family Foundation for supporting our work and Howard Gardner and Wendy Fischman for their thoughtful feedback.

\bibliography{custom}

%\appendix
%\label{sec:appendix}

%\section{Appendix}
%\begin{figure*}[t]
%\centering
%\resizebox{\textwidth}{!}{\input{figs/ToolWorkflow.tikz}}
%\caption{Example workflow with iterative loops for auditability.}
%\label{fig:audit-workflow}
%\end{figure*}

\end{document}